\documentclass[conference]{IEEEtran}
\IEEEoverridecommandlockouts
\usepackage{cite}
\usepackage{amsmath,amssymb,amsfonts}
\usepackage{algorithmic}
\usepackage{graphicx}
\usepackage{textcomp}
\usepackage{subcaption}
\usepackage{multicol}
\usepackage{xcolor}
\usepackage{comment}
\usepackage{multirow}
\usepackage[ruled,lined,linesnumbered]{algorithm2e}
\usepackage{algorithmic}
\usepackage{algorithm2e} 
\usepackage{balance}

\def\BibTeX{{\rm B\kern-.05em{\sc i\kern-.025em b}\kern-.08em
    T\kern-.1667em\lower.7ex\hbox{E}\kern-.125emX}}
\begin{document}

\title{\vspace{5mm} Cellular-enabled Collaborative Robots Planning and Operations for Search-and-Rescue Scenarios} 
\vspace{1 cm}
\author{
\IEEEauthorblockN{Arnau Romero\IEEEauthorrefmark{1}\IEEEauthorrefmark{3}\textsuperscript{\textsection}, Carmen Delgado\IEEEauthorrefmark{1}, Lanfranco Zanzi\IEEEauthorrefmark{2}, Ra\'ul Su\'arez\IEEEauthorrefmark{3}, Xavier~Costa-P\'erez\IEEEauthorrefmark{1}\IEEEauthorrefmark{2}\IEEEauthorrefmark{5}}
\vspace{1mm}
\IEEEauthorblockA{\IEEEauthorrefmark{1}AI-Driven Systems, i2CAT Foundation, Spain. Email: \{name.surname\}@i2cat.net,}
\IEEEauthorblockA{\IEEEauthorrefmark{2}NEC Laboratories Europe GmbH, Germany. Email:\{name.surname\}@neclab.eu,}
\IEEEauthorblockA{\IEEEauthorrefmark{3}{Polytechnic University of Catalonia, UPC)}, Spain. Email:\{name.surname\}@upc.edu,}
\IEEEauthorblockA{\IEEEauthorrefmark{5}Institut Català de Recerca i Estudis Avançats, ICREA, Spain. Email:\{name.surname\}@icrea.cat\\
\vspace{-6mm}
}}

\maketitle
\begingroup\renewcommand\thefootnote{\textsection}
\footnotetext{Corresponding Author.}
\endgroup
\begin{abstract}
Mission-critical operations, particularly in the context of Search-and-Rescue (SAR) and emergency response situations, demand optimal performance and efficiency from every component involved to maximize the success probability of such operations. In these settings, cellular-enabled collaborative robotic systems have emerged as invaluable assets, assisting first responders in several tasks, ranging from victim localization to hazardous area exploration.
However, a critical limitation in the deployment of cellular-enabled collaborative robots in SAR missions is their energy budget, primarily supplied by batteries, which directly impacts their task execution and mobility.
This paper tackles this problem, and proposes a search-and-rescue framework for cellular-enabled collaborative robots use cases that, taking as input the area size to be explored, the robots fleet size, their energy profile,  exploration rate required and target response time, finds the minimum number of robots able to meet the SAR mission goals and the path they should follow to explore the area. Our results,  i) show that first responders can rely on a SAR cellular-enabled robotics framework when planning mission-critical operations to take informed decisions with limited resources, and, ii) illustrate the number of robots versus explored area and response time trade-off depending on the type of robot: wheeled vs quadruped.
\end{abstract}


\section{Introduction}
\label{sec:Introduction}

In mission-critical Search-and-Rescue (SAR) operations, the fast response times to disasters and emergencies is paramount for saving lives. First responders teams tend to risk their lives in these situations. Such risks can be mitigated by using mobile robots for victim localization and exploration of hazardous areas\cite{SAR_UGR_2022}. Effective coordination of a multi-robot fleet is essential to meet the rapid response requirements of SAR operations, optimizing zone exploration and task allocation while avoiding redundancy~\cite{Cao2022}. This coordination can be achieved through a centralized task planner in an edge server~\cite{LNORM_2023} that collects feedback from the robots, maps the explored area and generates optimal path plans for each robot.

However, mobile robots face a significant challenge in the form of energy constraints since they normally rely solely on batteries. Increasing battery capacity would lead to added weight, resulting in higher mobility energy consumption and thus, a design trade-off~\cite{Albonico2021}. Extensive research has explored energy-aware strategies to enhance overall efficiency~\cite{zakharov2020energy}\cite{tang2020reinforcement}\cite{carabin2017review}~\cite{rappaport_2017}. 
These features, including battery charging decisions, can also be integrated into the decision-making algorithms of edge/cloud task planners.

The efficiency of SAR operations hinges on both their execution and prior planning. Effective strategic planning involves allocating and distributing equipment resources across the deployment area to significantly reduce mission response times. Factors such as the battery capacity of the robots, required number of robots, and the characteristics of the area are key to develop a plan that anticipates the mission demands. 

Our previous work~\cite{online_OROS} proposed to integrate the orchestration logic from the mobile network infrastructure and the robot domains in an \emph{online} manner, thus enabling information exchange between the robots and a centralized control-level task planner in real-time. 
Despite achieving promising results in mission efficiency, the outcomes of such approaches heavily depend on the initial assumptions and conditions considered. %

Given the importance of such mission planning decisions, in this work, we propose a novel robotic SAR framework that enhances state-of-the-art orchestration strategies for mobile collaborative robots by introducing a SAR mission planning phase. Specifically, we introduce a mission planning building block that takes into account readily available information to first responders such as the area to be explored and the number of robots available and, considering mission goals such as exploration rate and response time, provides informed decisions on the number of robots required for a mission.

\section{Related Work}
\label{sec:SoTA}

\begin{figure*}[t]
  \centering
\includegraphics[width=1\textwidth]{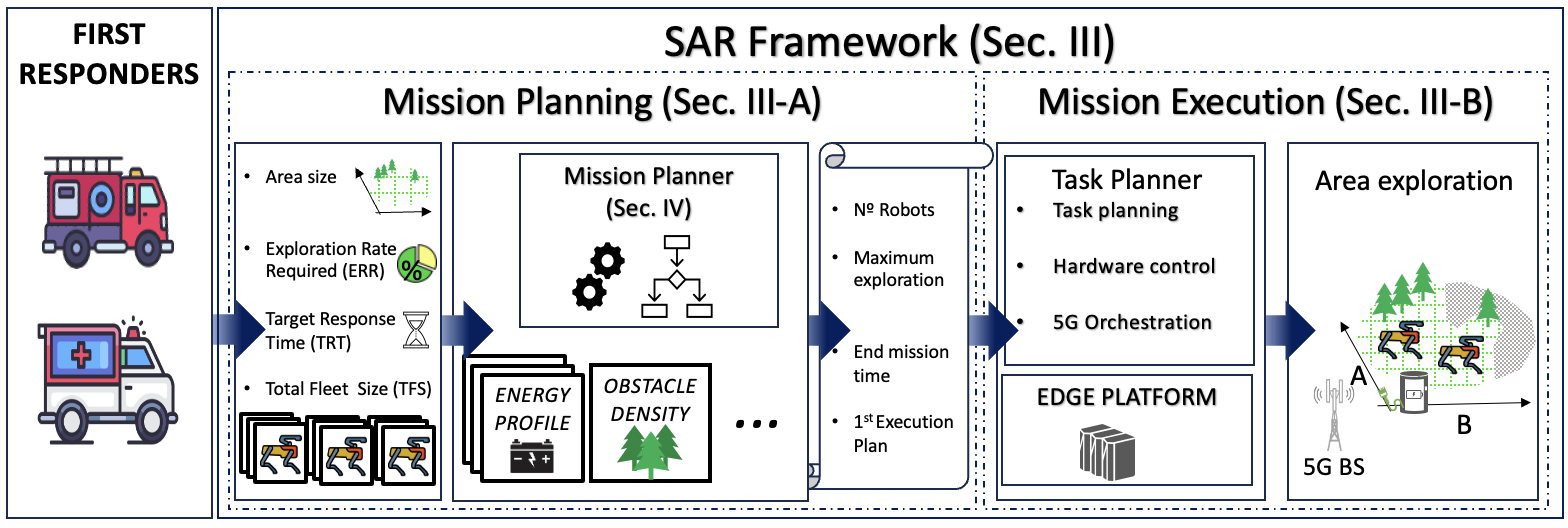}
  \caption{Cellular-enabled Collaborative Robotics Search-and-Rescue Framework.} 
\label{fig:Framework_fig}
\vspace{-3mm}
\end{figure*}

The adaptability and robustness of robot devices make them a valuable asset in hazardous environments, and it is no surprise that several works in the literature already investigated the adoption of mobile robots for SAR operations~\cite{SAR_review_2007}. In unstructured environments such as post-disaster areas, key metrics like response time and area coverage depend on multiple external factors, including the exploration strategy, the collaborative multi-robot system implementation, as well as robotic energetic and hardware resources~\cite{SAR_review_2020}.
Several works in the literature tackle these issues, but mostly in an independent manner. 
For example, the authors in ~\cite{tiwari2019} focus on the energy consumption estimation to extend the operational range. 
Similarly, 
in~\cite{maidana2020}, the authors propose to save energy by enabling/disabling certain robot hardware components, increasing the exploration efficiency and robot autonomy when revisiting already known areas. However, these works only focus on a single robot. 
Furthermore, other implementations such as computation offloading can significantly improve the energy savings as well\cite{chaari2022dynamic}.

In multi-robot SAR operations, the overall efficiency can be enhanced by an efficient robot coordination. In~\cite{LNORM_2023}, authors propose a centralized orchestration scheme for robot fleet path planning, leveraging edge computing and Wi-Fi technology for communication. 
Conversely, the authors in~\cite{offline_OROS} propose a joint 5G and robot orchestration logic that indicates optimal path planning of a robot fleet, as well as, robot hardware usage (i.e. sensors, communication peripherals) and battery charging priorities using a charging station.

The proposed architecture revolves around the possibility of using a dedicated 5G network by including a gNodeB 5G-NR base station in the SAR equipment deployment, which guarantees fast communications between the orchestrator and the robots. The results in this paper and in~\cite{online_OROS} denote that optimal performance in SAR operations is improved upon increasing the robot fleet size, while is also negatively affected by the exploration area obstacle density.

However, SAR performance can also be enhanced by focusing on the planning before mission execution. In~\cite{yan_2014}, the authors study the impact of robot fleet sizes on the area exploration performance. Results derived from a 3D frontier-based multi-robot collaborative framework denote a positive increase in exploration efficiency when increasing the fleet size, up to a certain limit. This justifies the existence of an optimal resource allocation point that depends on the specific target deployment area. Authors in~\cite{Yan_2015} also observed that the achieved exploration area and multi-robot performance depend on the initial position of robotic teams and need to be carefully considered during the initial planning phase.

None of the above works though have considered the energy aspects in their resource planning evaluation, or the impact the energy savings might have in the resource allocation previous to execution.

\section{Cellular-enabled Collaborative Robotics Search-And-Rescue Framework}
\label{sec:SARFramework}

 We consider scenarios where first responder teams leverage on a fleet of robots for SAR mission-critical operations in unknown areas.
 We assume the size of the exploration area is known (as it may be easily estimated), and that robots collaborate by making use of their cellular connectivity (4G/5G). 
 
 In order to optimize the first responders operation we design the  SAR framework depicted in Fig.~\ref{fig:Framework_fig}.
 It is composed of two main phases: the \emph{Mission Planning} phase and the \emph{Mission Execution} phase. In the following we describe them in detail.

\subsection{Mission Planning}
\label{subsec:PlanPhase}
The mission planning phase is designed as an offline step preceding mission deployment. In SAR missions, response time and equipment resources used are two key factors that determine their efficiency. In general, increasing the number of robots in collaborative scenarios tends to reduce the operation time. However, the number of robots available for missions is finite and their capabilities are limited by the battery capacity and consumption during operation. 

A \emph{mission planner} is thus needed to determine the minimum number of robots required for a mission, given an area to be explored and the available robot fleet size along with their characteristics (e.g., battery size, energy consumption, sensors, mobility), and considering the exploration rate required for the area and the maximum response time for a successful outcome. In Sec.~\ref{sec:Planner} we describe the designed mission planner in detail.

\subsection{Mission Execution}
\label{subsec:ExecPhase}
During the mission execution phase, the output of the mission planner is used as a starting point and updated during execution. In this phase, robots deployed in the field are expected to operate in coordination using navigation and exploration strategies to cover the target area. Additionally, using energy-aware strategies can also significantly increase exploration efficiency.
In our work, we assume that the responsibility of ensuring energy-aware path planning for the designated set of robots, achieved through the orchestration of their hardware and network resource utilization and task management, lies within the scope of an edge/cloud-based task planner.
The task planner is designed as a centralised high-level control entity that performs optimization decisions by performing hardware/software control, i.e., switch on/off peripherals and related drivers, as well as, cellular radio resource allocation. By integrating the capabilities of activating and deactivating sensors, communication peripherals and off-loading of computation processes, the task planner provides field robots with an energy-aware coordinated task plan upon area exploration.
Multiple examples of such task planners were discussed in Sec.~\ref{sec:SoTA}; in this work we will assume the usage of the task planner described in~\cite{online_OROS}.

\section{Mission Planner - Under the Hood}
\label{sec:Planner}

\begin{figure}[t]
      \centering
      \includegraphics[trim = 0cm 0cm 0cm 0cm , clip, width=1\columnwidth ]{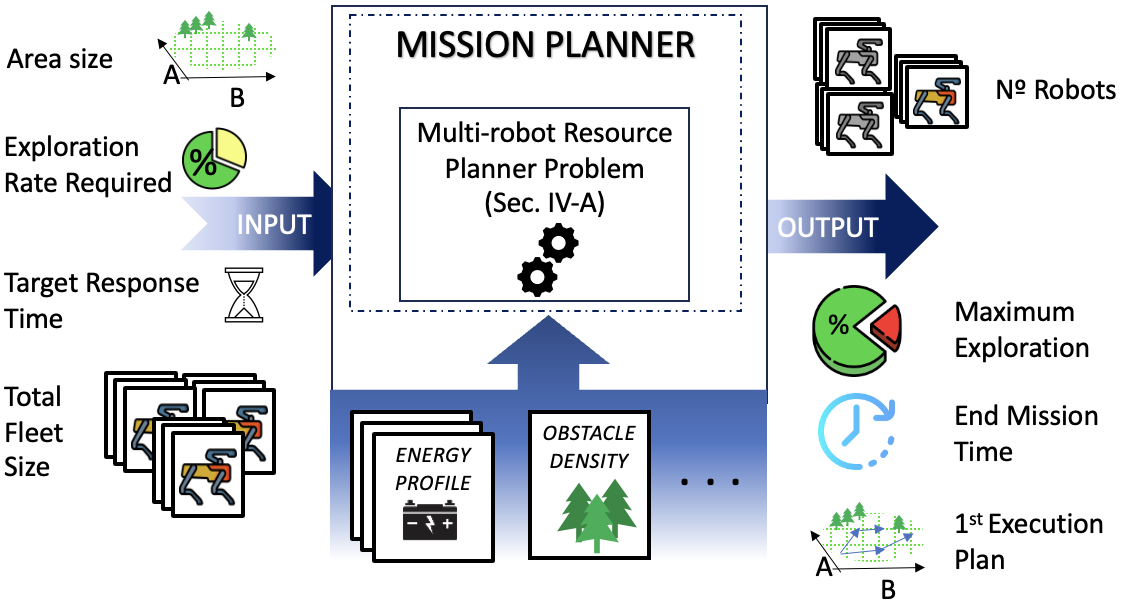}
      \caption{\small Overview of the Mission Planner. }
      \label{fig:SAR_Planner}
      \vspace{-3mm}
\end{figure}

In Fig.~\ref{fig:SAR_Planner} an overview of the Mission Planner is depicted. The input parameters are: i) exploration area size, ii) exploration rate required (ERR), i.e., the percentage of the whole area to be explored, iii) target response time (TRT), i.e., maximum time envisioned to complete a mission, and, iv) total robot fleet size (TFS). 
Then, using preloaded mission characteristics data (e.g., robots energy profiles, obstacle density) the mission planner launches a multi-robot resource planner optimizer to find the minimum fleet size for a given mission. As output, the Mission Planner provides: i) the number of robots to be used, ii) the expected exploration area, iii) the mission completion time, and, iv) an initial multi-robot path plan to perform during the mission execution.

Algorithm~\ref{alg:MP}, detailed below in this section, summarizes the Mission Planner implementation in pseudocode. As it can be observed, 
The Mission Planner iteratively evaluates a multi-robot resource planning problem considering an energy-aware optimization solution. At each iteration, we consider the usage of an increasing number of robots (with their corresponding energy profile and battery size) and determine the amount of time required to satisfy a predetermined ERR within a given TRT. The robot fleet size is increased by one at each iteration until either the ERR and TRT requirements are met or the total available fleet size is reached with no feasible solution.
A detailed description of the \emph{RP} optimization problem evaluated at each iteration is described next.

\begin{algorithm}[t]
\small
\SetKwInOut{Input}{Input}
\SetKwInOut{Output}{Output}
\SetKwInOut{Return}{return}
\SetKwInOut{Initialize}{Initialize}
\SetKwInOut{Procedure}{Procedure}
\Input{ $TRT, TFS, ERR$, $\mathcal{G}, m_{a,b,a',b'} $ \;}
\Procedure{}
      \While{ !solved}{
        UPDATE $\mathcal{R} \subset TFS$\;
        SOLVE \textit{RP} ( $\mathcal{T}, \mathcal{R}, ERR$) \;
        GET $d_{t},e_{t,a,b}, l_{r,t,a,b}\forall t \in \mathcal{T}$\;
        \If{$\sum d_{t} \leq TRT~OR~\mathcal{R} == TFS$}{
        solved = True\;
        } 
      }
      \Output{  $\mathcal{R},  d_{t}, l_{r,t,a,b}$\;}
\caption{Mission Planner}
\label{alg:MP}
\end{algorithm}

Hereafter, we present our assumptions, notation and problem formulation to model the multi-robot Resource Planner problem (RP), based on the adaptation of the problem formulation described in~\cite{offline_OROS}.

\textbf{{Input variables.}} Let us consider a discrete set of time instants $\mathcal{T} = \{t_1, \dots, t_{|\mathcal{T}|} \}$, and a set of robots $\mathcal{R}=\{r_1, \dots, r_{|\mathcal{R}|} \}$. Each robot is equipped with a battery characterized by a maximum capacity $B_{max}$,  $\forall r \in \mathcal{R}$, whose charging status $b_{r,t}$ varies over time depending on the robot activities and hardware usage. 
We assume the set of robots $\mathcal{R}$ to be deployed in an area of interest covered by mobile infrastructure for 5G connectivity. We consider the area of dimension $A \times B$ and discretize its 2D surface into a grid $\mathcal{G} =\{g_{a,b}, \forall (a,b) \in (A,B) \}$,  where each element $g_{a,b} \in \mathcal{G}$ needs to be explored.
We assume the same robot mobility approach described in \cite{offline_OROS}, where the motion energy consumption of a robot depends on a terrain-velocity constant $P_{move_{a,b,a',b'}}$. 
As mentioned before, robots exploit an existing mobile infrastructure for communications. We consider $P_{TXa,b}$ as a variable representing the energy consumed by a robot for transmitting data, and $P_{RX}$ for receiving data.
Finally, we collect the energy consumption derived by all the cameras and sensors, as well as their processing, in the variable~$P_{SEN}$.

\textbf{Decision variables.}
Let $d_{t}$ be a binary variable that determines whether there are still areas to be explored at a certain time $t \in \mathcal{T}$.  
To keep track of the multi-robot exploration, we introduce $e_{t,a,b}$ as a binary variable indicating if the area unit $g_{a,b}$ has already been explored at time $t \in \mathcal{T}$.  Additionally, $l_{r,t,a,b}$ is a binary decision variable to control the robot mobility. Its value gets positive if the robot $r$ is at position $g_{a,b}$ at time instant $t$.

\textbf{Constraints.} Since the algorithm needs to guarantee that the ERR is satisfied, we need to enforce that the total explored area in the last time step is at least the corresponding ERR (i.e., the percentage of the total area $|AB|$, now represented as $\kappa$). For this purpose, we include the following constraint:
\begin{equation}
    \sum_{(a,b) \in (A,B)} e_{t_f,a,b} \geq \kappa |AB| .
    \label{eq:exp_final}
\end{equation}

We ensure that each robot $r \in \mathcal{R}$ can only be in one place in every time instant  $t \in \mathcal{T}$:
\begin{equation}
	\label{eq:const7}
	\sum_{(a,b) \in (A,B)} l_{r,t,a,b} = 1  \quad \forall r \in \mathcal{R}, \forall t \in \mathcal{T},
\end{equation}
\noindent and with the following constraint we also ensure that the robots only move between neighbouring areas, or stay in the same position:
\begin{gather}
	l_{r,t+1,a,b} \leq l_{r,t,a,b} + l_{r,t,a-1,b} + l_{r,t,a+1,b} + l_{r,t,a,b-1} + \notag \\ 
	  l_{r,t,a,b+1} + 	l_{r,t,a-1,b-1} + l_{r,t,a+1,b+1} + l_{r,t,a-1,b+1} + \notag \\ 
 l_{r,t,a+1,b-1}  \quad \forall r \in \mathcal{R} , \forall t \in \mathcal{T}, \forall (a,b) \in (A,B).   \label{eq:const8} 
\end{gather}
In order to keep track of the exploration progress among multiple robots, if any robot $r \in \mathcal{R}$ visited 
an area unit $g_{a,b} \in (A, B)$ at some earlier time, or if it is exploring such area unit at the current time $t$, that area becomes explored at time $t$ and we update the variable $e_{t,a,b}$ accordingly.
\begin{equation}
	\label{eq:const3}
	e_{t,a,b} \leq 	e_{t-1,a,b} + \sum_{r \in \mathcal{R}}  l_{r,t,a,b}  \quad \forall t \in \mathcal{T}, \forall (a,b) \in (A,B),
\end{equation}
\begin{equation}
	\label{eq:const4}
	e_{t,a,b} \geq 		e_{t-1,a,b}  \quad \forall t \in \mathcal{T},    \forall (a,b) \in (A,B),  
\end{equation}
\begin{equation}
	\label{eq:const5}
	|\mathcal{R}| e_{t,a,b} \geq 		\sum_{r \in \mathcal{R}}  l_{r,t,a,b}   \quad \forall t \in \mathcal{T},    \forall (a,b) \in (A,B).  
\end{equation}

In order to update the decision variable $d_{t}$, according to the explored area at every time instant, we include the following constraint:
\begin{equation}
	\sum_{(a,b) \in (A,B)} e_{t,a,b} \geq \kappa (1 - d_t)  |AB|  \quad \forall t \in \mathcal{T}.
 \label{eq:act_d}
\end{equation}
Finally, as mentioned before, we assume the mobility consumption to be included in the constant $P_{move_{a,b,a',b'}}$ and mainly dependent on the robot velocity and terrain characteristics. For the robot communications, we assume that the robot can always receive data consuming $P_{RX}$. During data transmission, the consumed power depends on the distance to the base station (according to $P_{TX,a,b}$). If a robot has never been in an area unit, its sensors, camera, processing units and transmission elements should be active. However, to reduce the energy consumption, if the robot is in an already explored area, we consider the possibility of turning them off to save energy. Taking this into account, our algorithm updates the expected battery level $b_{r,t+1}$ as:
\begin{gather}
	b_{r,t+1} = b_{r,t}  - P_{RX} - \notag \\ 
	 \sum_{(a,b) \in (A,B)}  	\sum_{(a',b') \in (A,B)} l_{r,t,a,b} \times l_{r,t+1,a',b'} \times  P_{move_{a,b,a',b'}}  \notag \\ 
	- P_{SEN}  \times  \sum_{(a,b) \in (A,B)} (1 - e_{t,a,b}) \times l_{r,t+1,a,b} -  \label{eq:const9a}\\  
	\sum_{(a,b) \in (A,B)}   P_{TX,a,b} \times (1 - e_{t,a,b}) \times l_{r,t+1,a,b} \quad \forall t \in \mathcal{T} , \forall r \in \mathcal{R}. \notag
\end{gather}

\textbf{Objective.}
To increase the chances of detecting and assisting a target person in an unknown area it is necessary to minimize the time required to explore the target area:
\begin{equation}
	\label{eq:scala_obj_g2}
	\min  \sum_{t \in \mathcal{T}}  d_{t} 
\end{equation}
To sum up, the overall problem formulation of our multi-robot resource planner can be summarized as: 

\noindent \textbf{Problem}~\texttt{RP ($ \mathcal{T}, \mathcal{R}, \kappa$)} :
\label{prob:RP}
\begin{flalign}
  \quad\quad & \text{min} \sum_{t \in \mathcal{T}}  d_{t}  \nonumber & &\\
  \quad\quad & \text{subject to:} \nonumber\\
  \quad\quad & \quad\quad (\ref{eq:exp_final}) (\ref{eq:const7}) (\ref{eq:const8}) (\ref{eq:const3}) (\ref{eq:const4}) (\ref{eq:const5})  (\ref{eq:act_d}) (\ref{eq:const9a}); \nonumber
\end{flalign}

\section{Wheeled vs Quadruped Robots\\ Energy Profiling}
\label{sec:Energy Profiling}

\begin{figure}[t]
     \centering
     \begin{subfigure}{0.30\columnwidth}
         \centering
         \includegraphics[width=\textwidth]{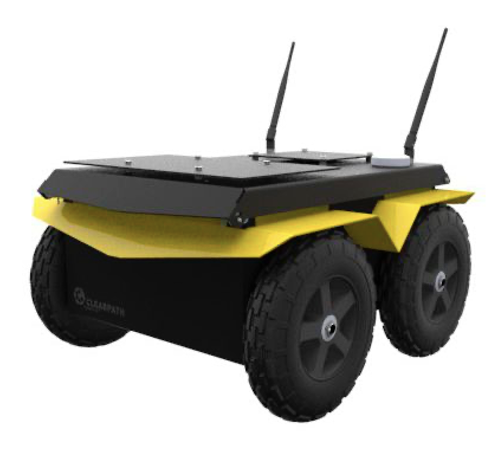}
         \caption{Wheeled robot}
         \vspace{3mm}
         \label{fig:wheeledRobot}
     \end{subfigure}
     \hspace{2mm}
     \begin{subfigure}{0.40\columnwidth}
         \centering
         \includegraphics[width=\textwidth]{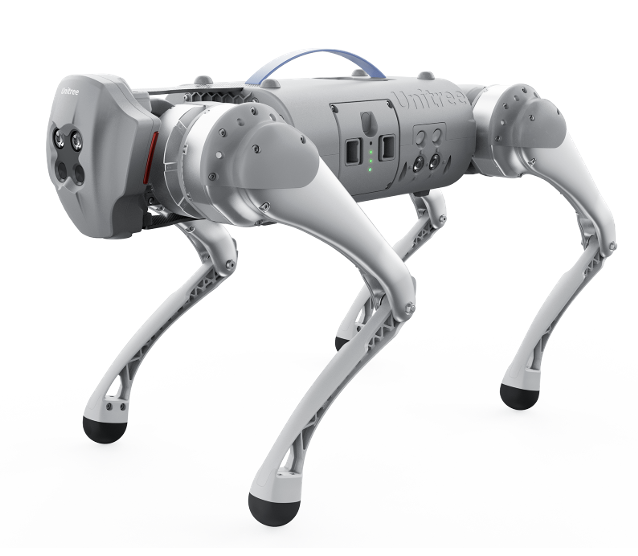}
         \caption{Quadruped robot}
         \vspace{3mm}
         \label{fig:quadrupedRobot}
     \end{subfigure}
     \vspace{-3mm}
     \caption{Robots evaluated in the SAR Framework.}
     \label{fig:resources}
     \vspace{-3mm}
\end{figure}

A key aspect to be considered for achieving an accurate mission plan is the energy profile of the robots. For this reason, in this section, we focus on analyzing the energy profile of two of the most common robot types used in SAR operations: wheeled and quadruped robots. Fig.~\ref{fig:resources} depicts examples of representative wheeled~\cite{Jackal} and quadruped robots~\cite{go1}.

Terrain-adaptability, motion speed and task-related energy consumption are key factors to be considered when choosing mobile robots for real-world scenarios, which should be carefully evaluated upon mission planning. While for cellular-enabled wheeled robots detailed energy profiling results already exist, e.g. \cite{online_OROS}, for quadruped robots no detailed energy profiling was found in the literature. Thus, to have a detailed model for our SAR framework of both wheeled and quadruped cellular-enabled robots, we acquired a Unitree GO1 EDU robot (see Fig. \ref{fig:quadrupedRobot}) and performed our own profiling. The results are summarized next. 

\subsection{Unitree GO1 Energy Profiling}
The Unitree GO1 is a quadruped robot equipped with a Raspberry Pi serving as the main CPU, supplemented by an array of three additional NVIDIA Jetson Nano units. Communication capabilities are facilitated through WiFi, Bluetooth, and a 4G QUECTEL chipset. The sensor suite of the robot comprises 5 pairs of cameras and 3 ultrasound sensors, with an additional feature being the inclusion of a 3D LiDAR that can be mounted on top the robot. Furthermore, Simultaneous Localization and Mapping (SLAM) using the LiDAR as well as human recognition through the camera feed can be performed. Notably, the Raspberry Pi perpetually powers the WiFi hotspot, while the cameras and ultrasound sensors rely on the Nano processors to which they are connected. Bluetooth, in contrast, remains in a dormant state until a new signal is received, rendering its power consumption negligible.

Table~\ref{tab:GO1_energy} presents a comprehensive breakdown of the obtained energy consumption associated with the Unitree GO1 EDU robot. Each row shows the results obtained when analysing the power consumption of independent robot components and motions. Tests have been performed averaging power consumption during a complete discharge of the 4500 mAh battery. For these measurements, \textit{unitree\_legged\_sdk} and \textit{unitree\_ros\_to\_real} ROS packages have been used to communicate through User Datagram Protocol to the controller, which publishes the robot state data, including the battery state\footnote{The data sets collected during the energy profiling measurements can be found at: github.com/armedarobotics/EnergyProfile-GO1.git}.

The table shows the results upon evaluating the consumption when enabling/disabling non-critical robot components (i.e., cellular communications, cameras, or processors). Power consumption has been determined by comparing it to a baseline of an idle standing robot state.
As can be seen, the main consumption is due to the use of SLAM techniques with a 3D LiDAR mounted on top of the robot. Similar consumption is observed when human recognition features (which uses the \textit{NVIDIA-AI-IOT trt\_pose}) are combined with the cameras.
The second section of the table denotes the results of the robot mobility tests. We have considered two possible idle states: up (standing) and down (laying). The results show that in a standing position, the robot consumes four times more energy than laying. Considering that it takes about 1~s for the robot to transition from up to down, and viceversa, the idle pose transition results denote that the robot can save energy by laying down in idle times longer than 2.87~s.
Four additional tests (driving straight at three different speeds and circling) have been performed moving the robot. The results, which are exclusively related to robot motion, denote that energy consumption tends to increase proportionally to the robot speed.

\begin{table}[t]
\caption{Power consumption breakdown of the GO1 robot}
\label{tab:GO1_energy}
\begin{tabular}{c|l|c|}
\cline{2-3}
\multicolumn{1}{l|}{}                             & \textbf{Consumption Element} & \multicolumn{1}{l|}{\textbf{Avg. Consumption (W)}} \\ \hline
\multicolumn{1}{|c|}{\multirow{4}{*}{\textbf{Components}}} & 4G Peripheral                & 15.77                                                 \\ \cline{2-3} 
\multicolumn{1}{|c|}{}                            & Cameras and Nano Proc.  & 19.25                                                 \\ \cline{2-3} 
\multicolumn{1}{|c|}{}                            & Human Recognition    & 29.38                                                 \\ \cline{2-3} 
\multicolumn{1}{|c|}{}                            & 3D LiDAR and SLAM     & 56.84                                                 \\ \hline\hline
\multicolumn{1}{|c|}{\multirow{8}{*}{\textbf{Mobility}}}   & Idle Down                    & 21.62                                                 \\ \cline{2-3} 
\multicolumn{1}{|c|}{}                            & Flex Down                    & 75.79                                                 \\ \cline{2-3} 
\multicolumn{1}{|c|}{}                            & Flex Up                      & 93.14                                                 \\ \cline{2-3} 
\multicolumn{1}{|c|}{}                            & Idle Up                      & 80.33                                                 \\ \cline{2-3} 
\multicolumn{1}{|c|}{}                            & Walking Circles 0.76 rad/s          & 73.86                                                \\ \cline{2-3} 
\multicolumn{1}{|c|}{}                            & Walking 0.5 m/s              & 53.26                                                \\ \cline{2-3} 
\multicolumn{1}{|c|}{}                            & Walking 1 m/s                & 108.86                                                \\ \cline{2-3} 
\multicolumn{1}{|c|}{}                            & Walking 2 m/s                & 211.22                                                \\ \hline
\end{tabular}
\vspace{-4mm}
\end{table}

\subsection{Wheeled vs Quadruped Energy Profiles}

Table~\ref{tab:GO1_OROS_energy} summarizes the energy profiling of both types of robots, taking the values from \cite{online_OROS} for the wheeled one and our own Unitree GO1 EDU profiling for the quadruped one. We compare the power consumption related to cellular communications for the reception and transmission of data, considering that both robots use similar technologies. As for sensing, consumption is related to the use of cameras, LiDAR sensor, SLAM processes and the processor. Robot inactivity is defined as idle state, and it is the minimum consumption robots have when performing no movement, nor using any particular hardware nor performing any action. As it can be observed, the quadruped robot has an overall higher power consumption than the wheeled robot. On the one hand, this is due to the quadruped robot having more consuming hardware components and processes than the wheeled robot. On the other hand, quadruped robots allow for more payload and are designed to explore more unstructured areas. In fact, upon evaluating the percentages of consumption rate, it is observed that it also consumes when standing in idle state. At the same time, wheeled robots normally have lower battery sizes than quadruped ones.

\begin{table}[t]
\caption{Power consumption comparison}
\label{tab:GO1_OROS_energy}
\resizebox{\columnwidth}{!}{\begin{tabular}{l|cc|cc|}
\cline{2-5}
\multicolumn{1}{c|}{}                             & \multicolumn{2}{c|}{\textbf{Avg. Consumption (W)}} & \multicolumn{2}{c|}{\textbf{Consumption rate  (\%)}} \\ \cline{2-5} 
\multicolumn{1}{c|}{}                             & \multicolumn{1}{c|}{Quadruped}        & Wheeled       & \multicolumn{1}{c|}{Quadruped}       & Wheeled       \\ \hline
\multicolumn{1}{|l|}{Cellular Reception}          & \multicolumn{1}{c|}{15.77}            & 4          & \multicolumn{1}{c|}{5.30\%}          & 13.97\%       \\ \hline
\multicolumn{1}{|l|}{Cellular Transmission}       & \multicolumn{1}{c|}{16.72}            & 4.95          & \multicolumn{1}{c|}{5.61\%}          & 17.28\%       \\ \hline
\multicolumn{1}{|l|}{Camera, LiDAR, Processor} & \multicolumn{1}{c|}{76.09}            & 12         & \multicolumn{1}{c|}{25.55\%}         & 41.90\%       \\ \hline
\multicolumn{1}{|l|}{Idle Up or Idle}             & \multicolumn{1}{c|}{80.33}            & 0.29          & \multicolumn{1}{c|}{26.98\%}         & 1.01\%        \\ \hline
\multicolumn{1}{|l|}{Motion 1 m/s}               & \multicolumn{1}{c|}{108.86}           & 7.40          & \multicolumn{1}{c|}{36.56\%}         & 25.84\%       \\ \hline
\multicolumn{1}{r|}{Total:}                       & \multicolumn{1}{c|}{297.77}           & 28.64   \\ \cline{2-3}       

\end{tabular}}
\vspace{-4mm}
\end{table}

\section{Performance Evaluation}
\label{sec:Evaluation}

In this section, we evaluate through simulations the performance of the cellular-enabled collaborative robotics SAR framework designed with a special focus on the Mission Planner building block described in Section \ref{sec:Planner} and considering both wheeled and quadruped robots with the energy profiles summarized in Sec.~\ref{sec:Energy Profiling}. 

\subsection{Evaluation Scenario Setup}

For the performance evaluation scenario setup, we consider two  exploration area sizes (50x50 $\mbox{m}^2$ and 500x500~$\mbox{m}^2$) to cover scenarios where the battery plays a negligible and a major role. Moreover, for the wheeled and quadruped robots we consider a battery maximum capacity of 72 kJ and 350 kJ, respectively, based on the specifications of each robot. 
Finally, as a first approximation to evaluate the trade-offs involved in the optimization problem, no obstacles and planar terrains are considered in the deployment scenarios to observe the impact of different fleet sizes in ideal conditions.

\begin{figure*}[ht]
     \centering
     \begin{subfigure}{0.40\textwidth}
         \centering
         \includegraphics[width=\columnwidth]{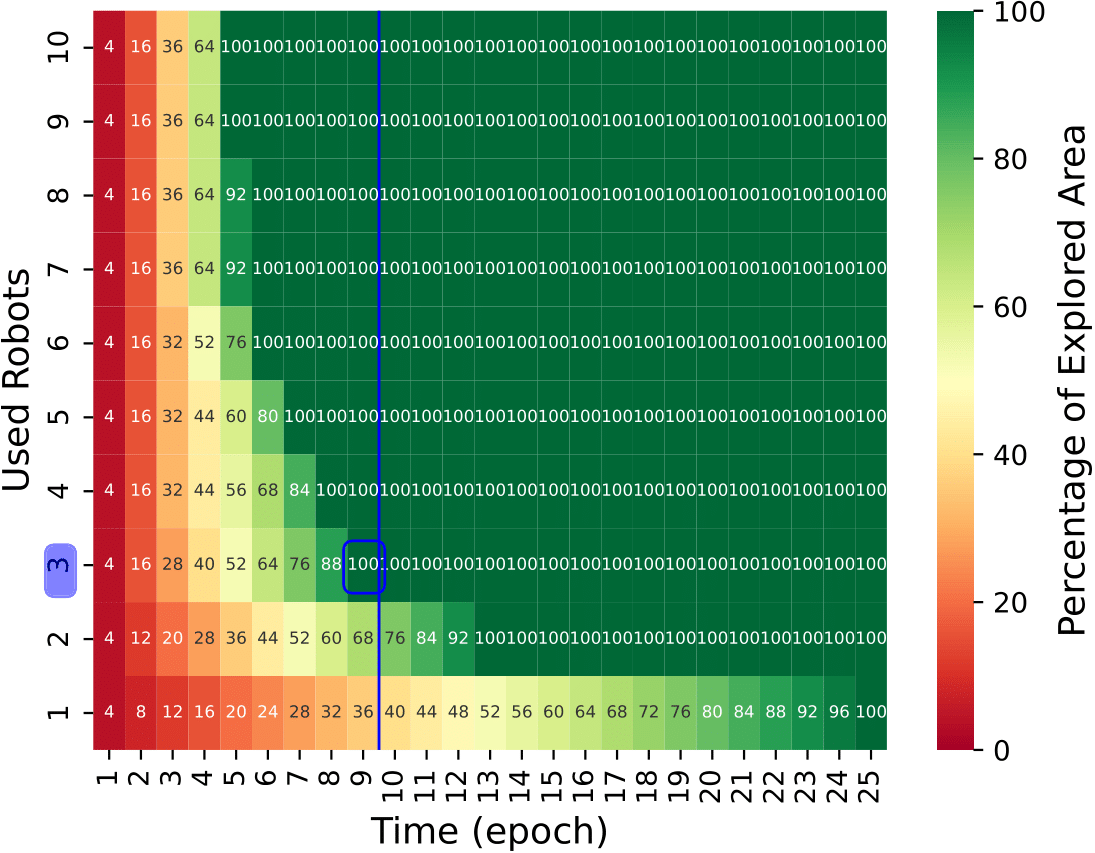}
         \caption{Wheeled robot}
         \vspace{1mm}
         \label{fig:wheeled}
     \end{subfigure}
     \hspace{3mm}
     \begin{subfigure}{0.40\textwidth}
         \centering
         \includegraphics[width=\columnwidth]{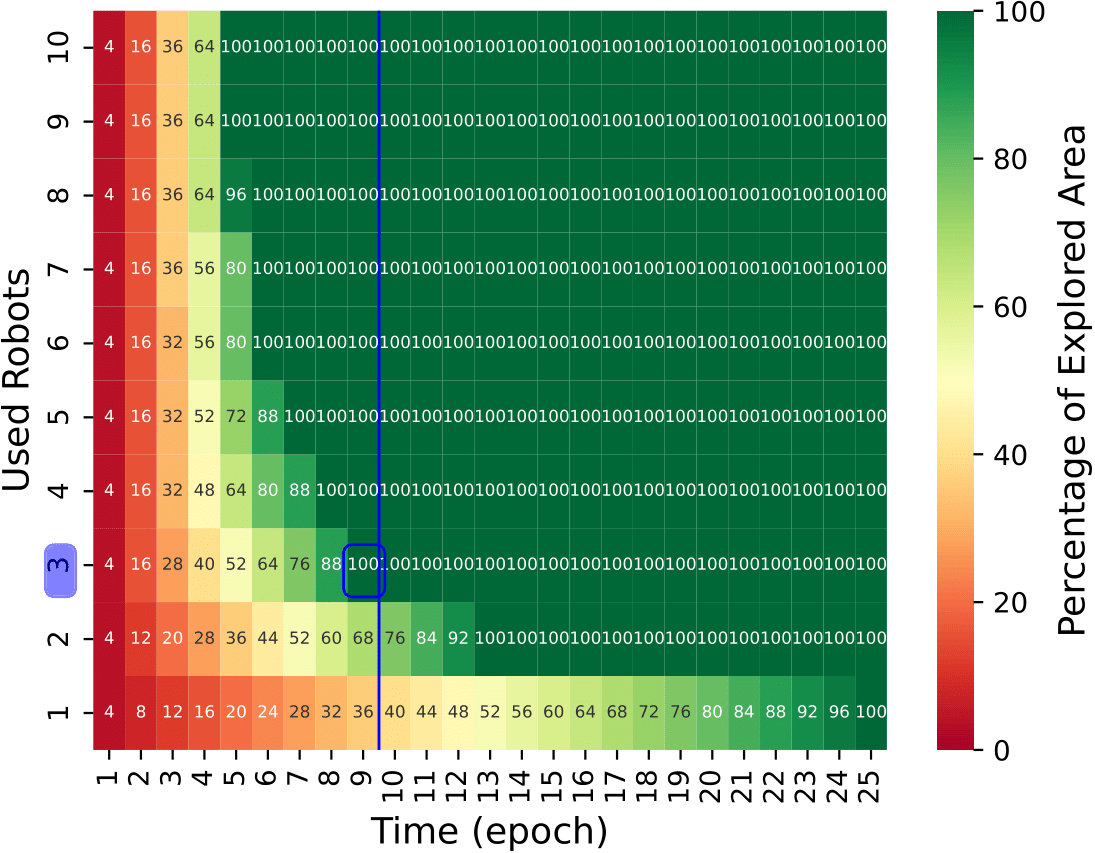}
         \caption{Quadruped robot}
         \vspace{1mm}
         \label{fig:quadruped}
     \end{subfigure}
     \vspace{-3mm}
     \caption{Percentage of Explored Area per Time Epoch for a 50x50 $\mbox{m}^2$ scenario. Wheeled versus Quadruped Robots.}
     \label{fig:comparative_legg_wheel}
     
\end{figure*}

\begin{figure*}[ht]
     \centering
     \begin{subfigure}{0.48\textwidth}
         \centering
         \includegraphics[width=\columnwidth]{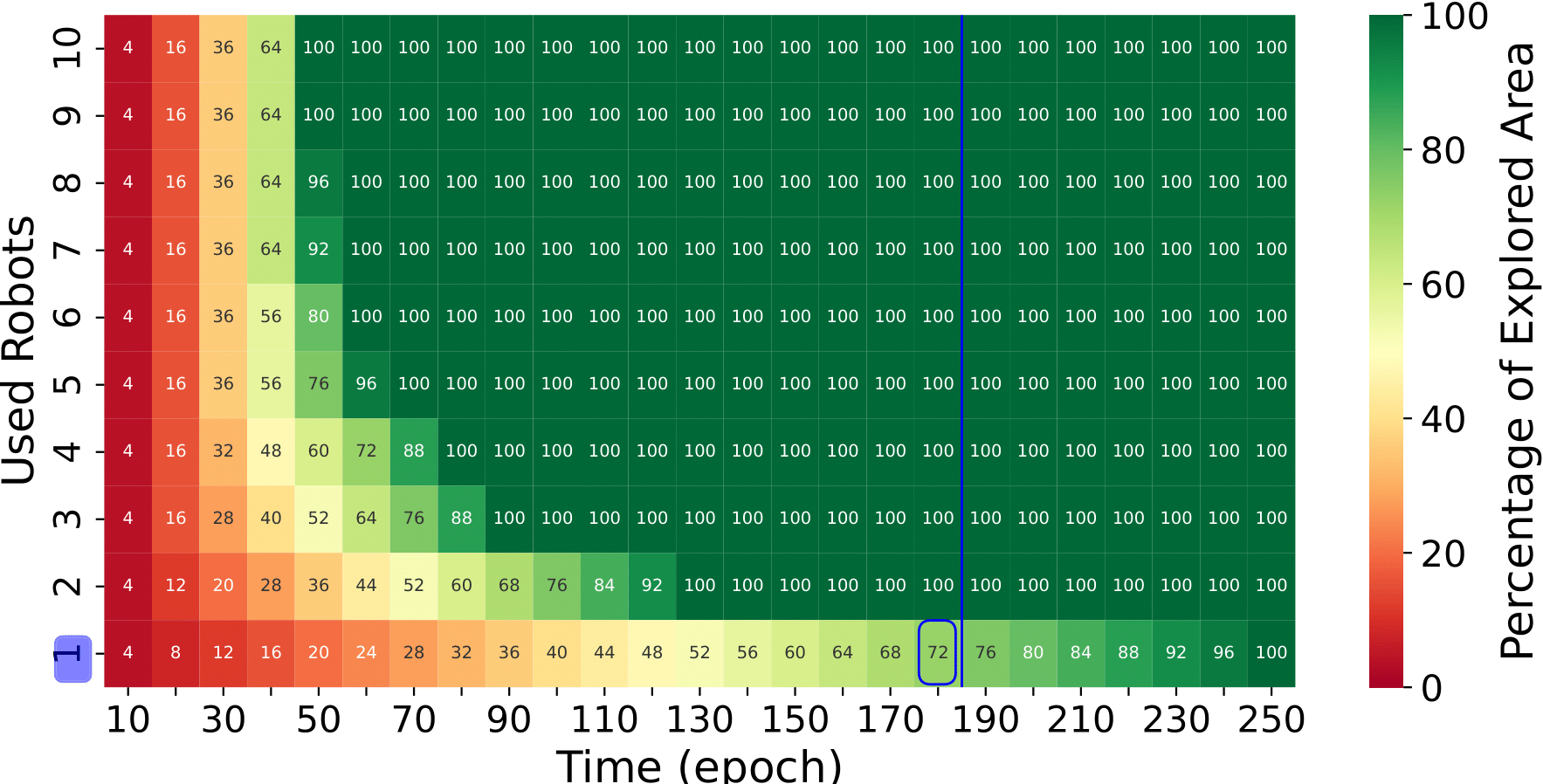}
         \caption{Wheeled robot}
         \vspace{2mm}
         \label{fig:wheeled_500}
     \end{subfigure}
     \hspace{2mm}
     \begin{subfigure}{0.48\textwidth}
         \centering
         \includegraphics[width=\columnwidth]{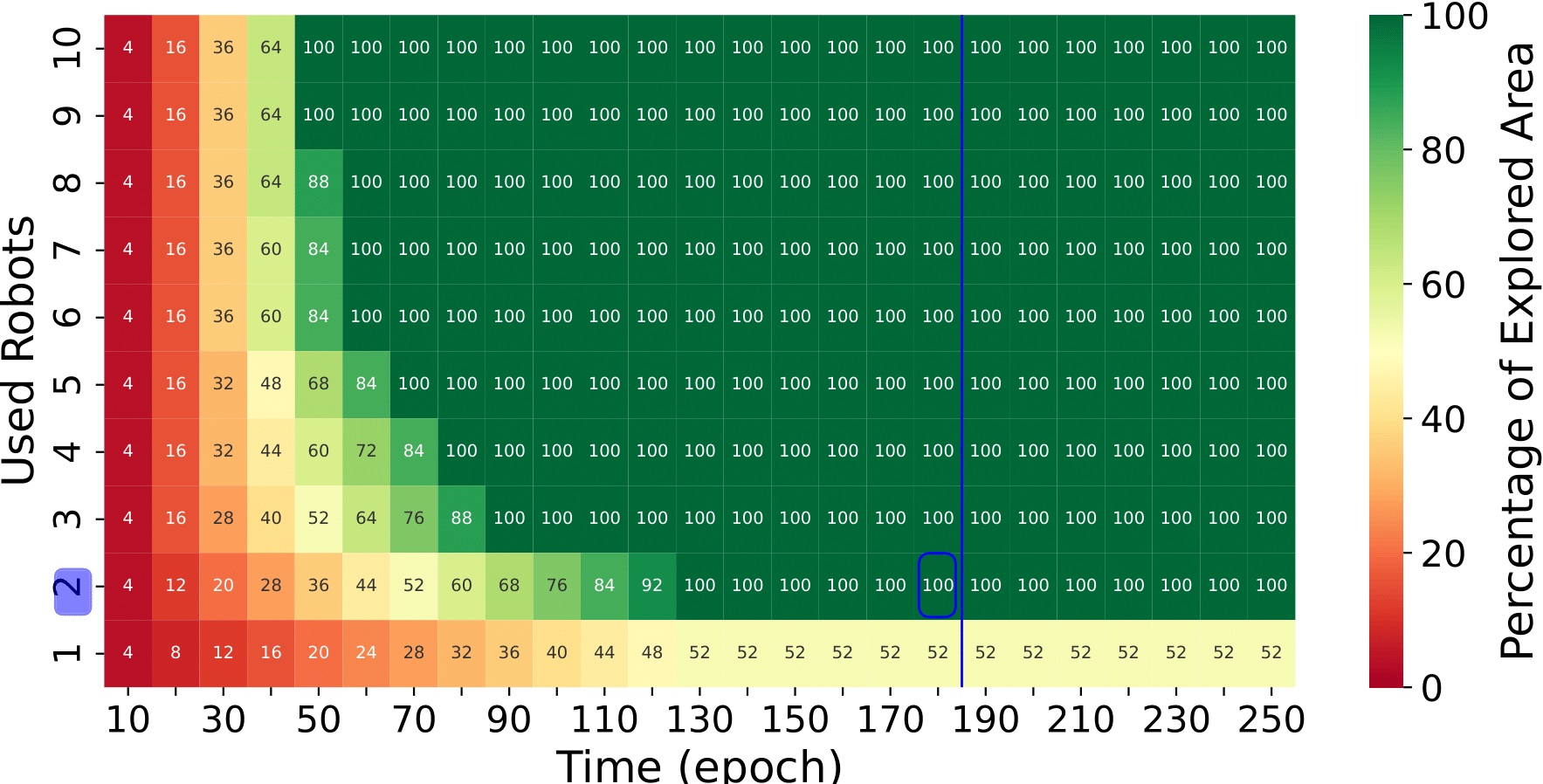}
         \caption{Quadruped robot}
         \vspace{2mm}
         \label{fig:quadruped_500}
     \end{subfigure}
     \vspace{-3mm}
     \caption{Percentage of Explored Area per Time Epoch for a 500x500 $\mbox{m}^2$ scenario. Wheeled versus Quadruped Robots.}
     \label{fig:comparative_legg_wheel_500}
     \vspace{-3mm}
\end{figure*}

\subsection{Mission Planning Evaluation}

In Fig.~\ref{fig:comparative_legg_wheel} we compare the performance of wheeled versus quadruped robot fleets in a 50x50 $\mbox{m}^2$ area moving both robots at 1 m/s. We consider a total fleet size of 10 robots, a required exploration rate of 70\% of the total area and a target response time up to 90~s. Note that each epoch is equivalent to 10~s in our experiments. 
The results obtained with both types of robots are similar, as the battery capacity is sufficient to cover the totality of the area at the given 1 m/s speed. In this scenario, the Mission Planner outputs a minimal fleet size of three robots to satisfy the target conditions, for both types of robots.

In Fig.~\ref{fig:comparative_legg_wheel_500} the Mission Planner has been used to evaluate the impact of a ten times larger area (500x500 $\mbox{m}^2$).
As in the previous case, we consider a total fleet size of 10 robots, a required exploration rate of 70\% of the total area, a target response time up to 180 epochs given the larger size of the scenario and a moving speed of 1 m/s. 
The results in this case differ between the wheeled and quadruped robots as expected since the differences in the energy consumption between the robot types become visible.
Although quadruped robots require more energy, this is offset by their larger battery size. However, they still have a greater impact on the battery capacity compared to wheeled robots. As a result, the Mission Planner suggests that while one wheeled robot would suffice for the mission, two quadruped robots would be necessary under similar conditions.

\section{Conclusions and Future Works}
\label{sec:Conclusions}

In mission-critical operations, the role of cellular-enabled collaborative robot fleets in augmenting the search-and-rescue capabilities of first responders is crucial. 

In this paper, we proposed a novel SAR framework for cellular-enabled collaborative robotics mission planning that, taking as input information readily available (exploration area, fleet size, energy profile, target exploration rate and target response time), allows first responders to take informed decisions about the number of robots needed to successfully complete a mission. Moreover, our results illustrated the trade-off involved when considering different types of robots (wheeled vs quadruped) with respect to the number of robots, explored area and response time.

Future work will consider expanding our SAR framework to further consider larger scale scenarios (e.g., in terms of larger areas, number and heterogeneity of robots, terrain diversity, obstacles) and input parameters available (e.g., detailed surface information, higher granularity of robot energy and control). In such cases, the problem complexity might get increasingly daunting but if analytical and/or machine learning solutions can be applied to make them feasible, better-informed decisions will be enabled.

\section*{Acknowledgment}
The research leading to these results has been supported by the Spanish Ministry of Economic Affairs and Digital Transformation and the European Union - NextGeneration EU, in the framework of the Recovery Plan, Transformation and Resilience (PRTR) (Call UNICO I+D 5G 2021, ref. number TSI-063000-2021-6), by the CERCA Programme from the Generalitat de Catalunya, by the European Union's H2020 5G ERA Project (grant no. 101016681), and by Spanish Ministry of Science and Innovation through the project PID2020-114819GB-I00.

\balance
\bibliographystyle{IEEEtran}
\bibliography{biblio}

\begin{thebibliography}{10}
\providecommand{\url}[1]{#1}
\csname url@samestyle\endcsname
\providecommand{\newblock}{\relax}
\providecommand{\bibinfo}[2]{#2}
\providecommand{\BIBentrySTDinterwordspacing}{\spaceskip=0pt\relax}
\providecommand{\BIBentryALTinterwordstretchfactor}{4}
\providecommand{\BIBentryALTinterwordspacing}{\spaceskip=\fontdimen2\font plus
\BIBentryALTinterwordstretchfactor\fontdimen3\font minus \fontdimen4\font\relax}
\providecommand{\BIBforeignlanguage}[2]{{%
\expandafter\ifx\csname l@#1\endcsname\relax
\typeout{** WARNING: IEEEtran.bst: No hyphenation pattern has been}%
\typeout{** loaded for the language `#1'. Using the pattern for}%
\typeout{** the default language instead.}%
\else
\language=\csname l@#1\endcsname
\fi
#2}}
\providecommand{\BIBdecl}{\relax}
\BIBdecl

\bibitem{SAR_UGR_2022}
D.~Huamanchahua, K.~Aubert, M.~Rivas, E.~Guerrero, L.~Kodaka, and D.~Guevara, ``Land-mobile robots for rescue and search: A technological and systematic review,'' in \emph{2022 IEEE International IOT, Electronics and Mechatronics Conference (IEMTRONICS)}, 2022, pp. 1--6.

\bibitem{Cao2022}
C.~Cao, H.~Zhu, F.~Yang, Y.~Xia, H.~Choset, J.~Oh, and J.~Zhang, ``Autonomous exploration development environment and the planning algorithms,'' in \emph{2022 International Conference on Robotics and Automation (ICRA)}, 2022, pp. 8921--8928.

\bibitem{LNORM_2023}
S.~Mohanti, D.~Roy, M.~Eisen, D.~Cavalcanti, and K.~Chowdhury, ``L-norm: Learning and network orchestration at the edge for robot connectivity and mobility in factory floor environments,'' \emph{IEEE Transactions on Mobile Computing}, pp. 1--16, 2023.

\bibitem{Albonico2021}
M.~Albonico, I.~Malavolta, G.~Pinto, E.~Guzman, K.~Chinnappan, and P.~Lago, ``{Mining Energy-Related Practices in Robotics Software},'' in \emph{Mining Software Repositories Conference (MSR)}, May 2021.

\bibitem{zakharov2020energy}
K.~Zakharov, A.~Saveliev, and O.~Sivchenko, ``Energy-efficient path planning algorithm on three-dimensional large-scale terrain maps for mobile robots,'' in \emph{Interactive Collaborative Robotics: 5th International Conference, ICR 2020, St Petersburg, Russia, October 7-9, 2020, Proceedings 5}.\hskip 1em plus 0.5em minus 0.4em\relax Springer, 2020, pp. 319--330.

\bibitem{tang2020reinforcement}
G.~Tang, N.~Kumar, and K.~P. Michmizos, ``Reinforcement co-learning of deep and spiking neural networks for energy-efficient mapless navigation with neuromorphic hardware,'' in \emph{2020 IEEE/RSJ International Conference on Intelligent Robots and Systems (IROS)}.\hskip 1em plus 0.5em minus 0.4em\relax IEEE, 2020, pp. 6090--6097.

\bibitem{carabin2017review}
G.~Carabin, E.~Wehrle, and R.~Vidoni, ``A review on energy-saving optimization methods for robotic and automatic systems,'' \emph{Robotics}, vol.~6, no.~4, p.~39, 2017.

\bibitem{rappaport_2017}
M.~Rappaport and C.~Bettstetter, ``Coordinated recharging of mobile robots during exploration,'' in \emph{2017 IEEE/RSJ International Conference on Intelligent Robots and Systems (IROS)}, 2017, pp. 6809--6816.

\bibitem{online_OROS}
A.~Romero, C.~Delgado, L.~Zanzi, X.~Li, and X.~Costa-Pérez, ``{OROS: Online Operation and Orchestration of Collaborative Robots using 5G},'' \emph{IEEE Transactions on Network and Service Management}, pp. 1--1, 2023.

\bibitem{SAR_review_2007}
J.~Liu, Y.~Wang, B.~Li, and S.~Ma, ``Current research, key performances and future development of search and rescue robots,'' \emph{Frontiers of Mechanical Engineering in China}, vol.~2, pp. 404--416, 2007.

\bibitem{SAR_review_2020}
J.~P. Queralta, J.~Taipalmaa, B.~Can~Pullinen, V.~K. Sarker, T.~Nguyen~Gia, H.~Tenhunen, M.~Gabbouj, J.~Raitoharju, and T.~Westerlund, ``Collaborative multi-robot search and rescue: Planning, coordination, perception, and active vision,'' \emph{IEEE Access}, vol.~8, pp. 191\,617--191\,643, 2020.

\bibitem{tiwari2019}
K.~Tiwari, X.~Xiao, A.~Malik, and N.~Y. Chong, ``A unified framework for operational range estimation of mobile robots operating on a single discharge to avoid complete immobilization,'' \emph{Mechatronics}, vol.~57, pp. 173--187, 2019.

\bibitem{maidana2020}
R.~G. Maidana, R.~Granada, D.~Jurak, M.~Magnaguagno, F.~Meneguzzi, and A.~Amory, ``Energy-aware path planning for autonomous mobile robot navigation,'' in \emph{International FLAIRS Conference}, 2020, pp. 362--367.

\bibitem{chaari2022dynamic}
R.~Cha{\^a}ri, O.~Cheikhrouhou, A.~Koub{\^a}a, H.~Youssef, and T.~N. Gia, ``Dynamic computation offloading for ground and flying robots: Taxonomy, state of art, and future directions,'' \emph{Computer Science Review}, vol.~45, p. 100488, 2022.

\bibitem{offline_OROS}
C.~Delgado, L.~Zanzi, X.~Li, and X.~Costa-Pérez, ``{OROS: Orchestrating ROS-driven Collaborative Connected Robots in Mission-Critical Operations},'' in \emph{IEEE International Symposium on a World of Wireless, Mobile and Multimedia Networks (WoWMoM)}, 2022, pp. 147--156.

\bibitem{yan_2014}
Z.~Yan, L.~Fabresse, J.~Laval, and N.~Bouraqadi, ``Team size optimization for multi-robot exploration,'' in \emph{Simulation, Modeling, and Programming for Autonomous Robots: 4th International Conference, SIMPAR 2014, Bergamo, Italy, October 20-23, 2014. Proceedings 4}.\hskip 1em plus 0.5em minus 0.4em\relax Springer, 2014, pp. 438--449.

\bibitem{Yan_2015}
Z.~Yan \emph{et~al.}, ``Metrics for performance benchmarking of multi-robot exploration,'' in \emph{2015 IEEE/RSJ International Conference on Intelligent Robots and Systems (IROS)}.\hskip 1em plus 0.5em minus 0.4em\relax IEEE, 2015, pp. 3407--3414.

\bibitem{Jackal}
\BIBentryALTinterwordspacing
{Clearpath Robotics Inc}. (2016) {Jackal UGV - Small Weatherproof Robot - Clearpath}. [Online]. Available: \url{https://clearpathrobotics.com/jackal-small-unmanned-ground-vehicle/}
\BIBentrySTDinterwordspacing

\bibitem{go1}
\BIBentryALTinterwordspacing
{Unitree Robotics}. (2021) {Unitree GO1 - UnitreeRobotics}. [Online]. Available: \url{https://shop.unitree.com/products/unitreeyushutechnologydog-artificial-intelligence-companion-bionic-companion-intelligent-robot-go1-quadruped-robot-dog}
\BIBentrySTDinterwordspacing

\end{thebibliography}

\end{document}